%% file: main_arxiv.tex
\documentclass[10pt,twocolumn,letterpaper]{article}

\usepackage{cvpr}              

\usepackage{graphicx}
\usepackage{amsmath}
\usepackage{amssymb}
\usepackage{booktabs}
\usepackage{pifont}
\usepackage{colortbl}
\usepackage{xcolor}
\definecolor{LGray}{gray}{0.97}
\definecolor{LGray1}{gray}{0.9}
\definecolor{darkgreen}{rgb}{0.0, 0.5, 0.0}
\usepackage{amssymb}
\usepackage{graphicx}
\usepackage{array}
\usepackage{makecell}
\usepackage{tabularx}
\usepackage[most]{tcolorbox} 
\usepackage{microtype}
\tcbuselibrary{listings}
\usepackage{listings}
\usepackage{cuted} 
\newcolumntype{C}[1]{>{\centering\arraybackslash}m{#1}}
\newcolumntype{L}[1]{>{\raggedright\arraybackslash}p{#1}}
\newcolumntype{Y}{>{\raggedright\arraybackslash}X}
\renewcommand{\arraystretch}{1.18}    
\definecolor{cvprblue}{rgb}{0.21,0.49,0.74}
\usepackage[pagebackref,breaklinks,colorlinks]{hyperref}
\hypersetup{
  linkcolor=cvprblue,
  citecolor=cvprblue,
  urlcolor=cvprblue
}
\usepackage[capitalize]{cleveref}
\crefname{section}{Sec.}{Secs.}
\Crefname{section}{Section}{Sections}
\Crefname{table}{Table}{Tables}
\crefname{table}{Tab.}{Tabs.}

\def\cvprPaperID{*****} 
\def\confName{CVPR}
\def\confYear{2022}

\begin{document}

\title{\LARGE \bf
How Good are Foundation Models in Step-by-Step Embodied Reasoning?
}

\author{
    \begin{minipage}[t]{\textwidth}
        \centering
         {Dinura Dissanayake}\textsuperscript{1},
         {Ahmed Heakl}\textsuperscript{1},
         {Omkar Thawakar}\textsuperscript{1},
         {Noor Ahsan}\textsuperscript{1},
         {Ritesh Thawkar}\textsuperscript{1},
         {Ketan More}\textsuperscript{1},
         {Jean Lahoud}\textsuperscript{1},
         {Rao  Anwer}\textsuperscript{1},
         {Hisham Cholakkal}\textsuperscript{1},
         {Ivan Laptev}\textsuperscript{1},
         {Fahad Khan}\textsuperscript{1, 2},
         {Salman Khan}\textsuperscript{1, 3}\\ [1em]
        \normalsize{
            \textsuperscript{1}Mohamed bin Zayed University of AI, 
            \textsuperscript{2}Linköping University,
            \textsuperscript{3}Australian National University 
            }
        \end{minipage}
    \vspace{-1em}
}
\maketitle

\input{sec/0_abstract}

\input{sec/1_intro}

\input{sec/2_related_works}

\input{sec/3_benchmark_curation}

\input{sec/4_evaluation_framework}

\input{sec/5_benchmarking}

\input{sec/6_conclution}










{\small
\bibliographystyle{ieee_fullname}
\bibliography{main_arxiv}
}

\end{document}

%% file: sec/0_abstract.tex
\begin{abstract}
Embodied agents operating in the physical world must make decisions that are not only effective but also safe, spatially coherent, and grounded in context.
While recent advances in large multimodal models (LMMs) have shown promising capabilities in visual understanding and language generation, their ability to perform structured reasoning for real-world embodied tasks remains underexplored.
In this work, we aim to understand how well foundation models can perform step-by-step reasoning in embodied environments. To this end, we propose the Foundation Model Embodied Reasoning (FoMER) benchmark, designed to evaluate the reasoning capabilities of LMMs in complex embodied decision-making scenarios.
Our benchmark spans a diverse set of tasks that require agents to interpret multimodal observations, reason about physical constraints and safety, and generate valid next actions in natural language.
We present \textit{(i)} a large-scale, curated suite of embodied reasoning tasks, \textit{(ii)} a novel evaluation framework that disentangles perceptual grounding from action reasoning, and \textit{(iii)} empirical analysis of several leading LMMs under this setting.
Our benchmark includes over 1.1k samples with detailed step-by-step reasoning across 10 tasks and 8 embodiments, covering three different robot types. 
Our results highlight both the potential and current limitations of LMMs in embodied reasoning, pointing towards key challenges and opportunities for future research in robot intelligence. 
Project page: \href{https://mbzuai-oryx.github.io/FoMER-Bench/}{https://mbzuai-oryx.github.io/FoMER-Bench/} 
\end{abstract}

%% file: sec/1_intro.tex
\section{Introduction}
Embodied agents, including robots and virtual agents that perceive and act within the physical world, are commonly deployed in complex, unstructured environments. 
Whether navigating cluttered homes, manipulating objects in a kitchen, or assisting humans in dynamic settings, these agents must target effective decision-making.
To achieve an intelligent level of decision-making, the embodied agents require reasoning: the ability to understand task goals, assess spatial relationships among objects and surfaces, and account for safety and physical feasibility before taking an action.

\begin{figure}[!t]
    \centering
    \includegraphics[width=1\linewidth]{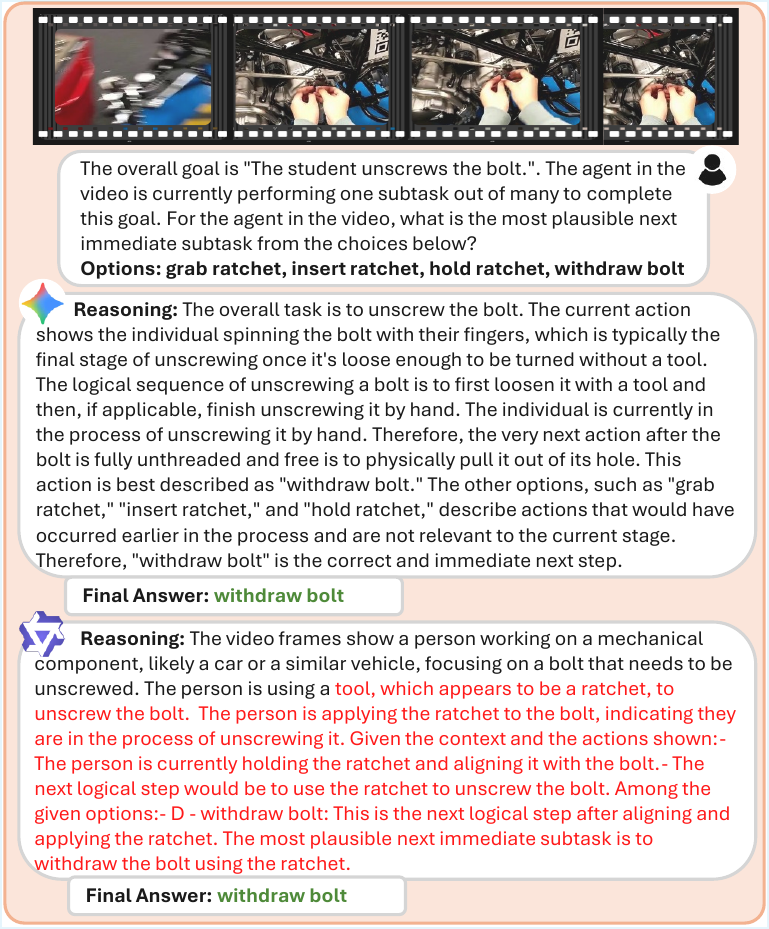}
    \caption{
    Example illustrating the final answer and step-by-step reasoning from Gemini (\includegraphics[height=1em]{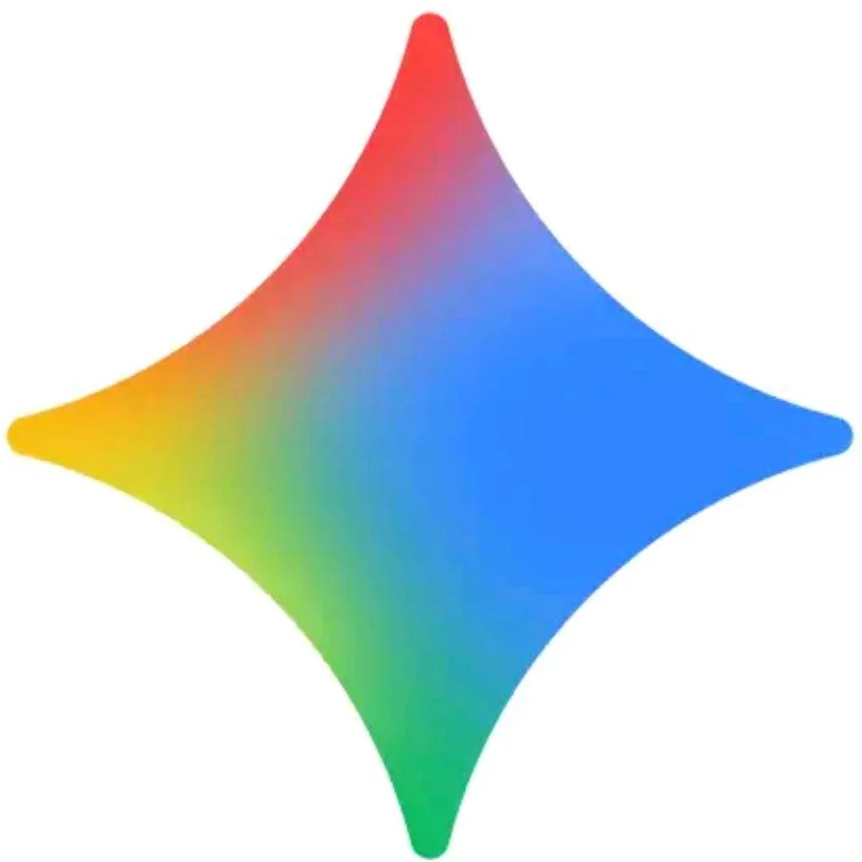}) \cite{comanici2025gemini25} and Qwen (\includegraphics[height=1em]{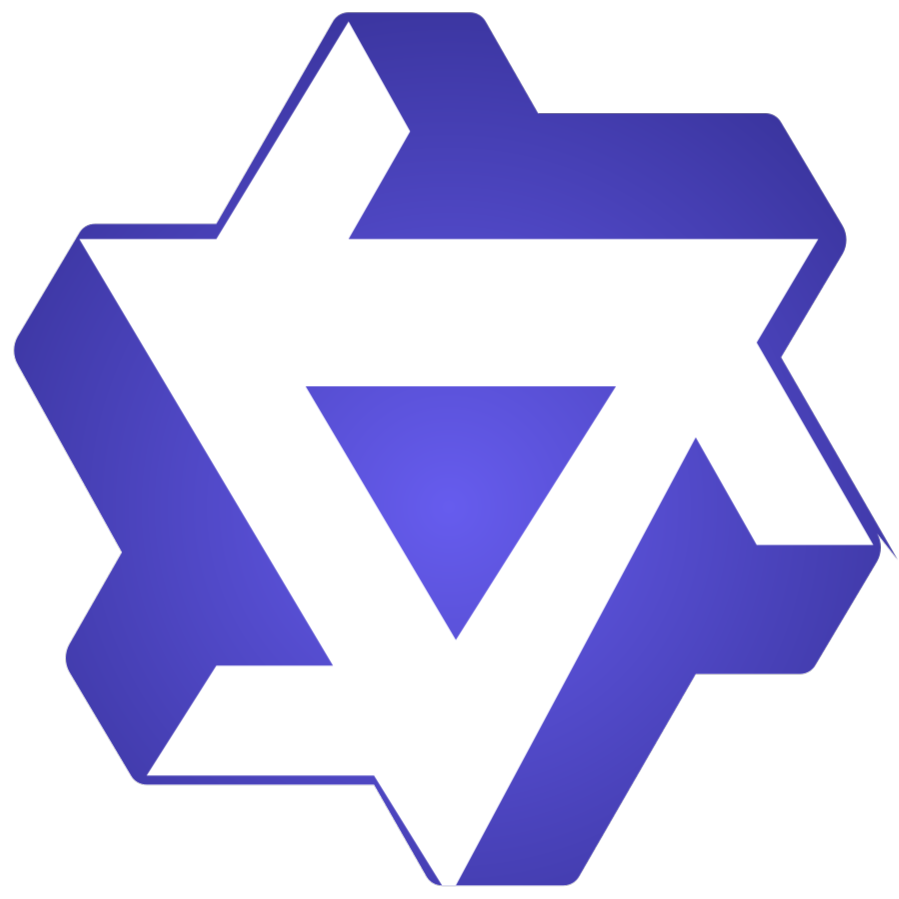}) \cite{Qwen2.5-VL} for a given video and text prompt. While both models correctly identify the action as ``withdraw bolt," the reasoning they provide differs significantly. This underscores the importance of evaluating the final answer as well as the underlying reasoning.}
    \label{fig:reasoning_trail}
\end{figure}

\begin{table*}[!t] 
\input{Tables/Benchmark_Comparison}
\end{table*}

While classical robotics addresses planning and control through explicit models of geometry and physics, the emergence of large multimodal models (LMMs) opens a new direction.
These models, which integrate vision and language through large-scale pretraining, have demonstrated impressive capabilities in visual understanding and language generation.
Prior work has applied LMMs or vision-language models to embodied contexts\cite{rt22023arxiv, kim24openvla}.
However, their potential for supporting complex decision-making, especially based on reasoning, remains largely underexplored.

On the other hand, outside of the field of embodied agents, several approaches have been presented to study and benchmark reasoning in LMMs \cite{thawakar-etal-2025-llamav}. 
These approaches often target general reasoning, such as commonsense, temporal, and abstract logic, among others.
Nevertheless, these works are typically detached from the embodied setting and fail to capture reasoning types that are crucial for physical interaction, such as spatial alignment, object affordance, and safety awareness.
As a result, there is a lack of clear understanding of whether and how LMMs can reason effectively in the context of embodied tasks.
%
This gap is particularly evident when examining how LLMs generate their final answers. Figure \ref{fig:reasoning_trail} shows how a model can arrive at the right conclusion through faulty reasoning. In this example, both Gemini 2.5 Pro \cite{comanici2025gemini25} and Qwen 2.5-VL \cite{Qwen2.5-VL} reach the same final answer but follow completely different reasoning paths. This highlights the need to evaluate not only the final answers, but also the reasoning trails behind them.

In this work, we introduce the Foundation Model Embodied Reasoning benchmark (FoMER), specifically designed to evaluate embodied reasoning in LLMs.
Our benchmark comprises a diverse set of scenarios in which an agent must respond to prompts in natural language, grounded in both visual observations and a task description.
We carefully annotate each scenario with a valid action response, alongside fine-grained reasoning labels that cover spatial relations, safety constraints, and task alignment.
When compared to existing embodied reasoning benchmarks (Table \ref{tab:methods_comparison}), our benchmark offers the largest coverage, both in the question diversity and in reasoning diversity.

%
To assess a model's performance, we propose an evaluation framework that measures not only the accuracy of the selected action but also the quality of the underlying reasoning, as judged against human-annotated reasoning.
This enables us to check cases where a model chooses responses based on guesses, rather than truly understanding the underlying decision logic. 

In summary, our contributions are as follows:
\begin{itemize}
    \item We present a large-scale benchmark for evaluating embodied reasoning in LMMs, spanning more than 1k scenarios with detailed reasoning annotations.
    \item We propose an evaluation framework that emphasizes grounded reasoning, introducing new metrics that capture both action validity and reasoning correctness.
    \item We benchmark several state-of-the-art LMMs under this framework, highlighting strengths and limitations in their ability to reason about real-world embodied tasks.
    
\end{itemize}

%% file: Tables/Benchmark_Comparison.tex
\centering
    \setlength{\tabcolsep}{4pt}
     \caption{A comparison of various Embodied and Physical AI benchmarks. We summarize key features across benchmarks, including input modalities, question formats, presence of step-by-step reasoning trails, number of annotated questions, annotation methods, diversity of tasks and embodiments, and the types of robots involved. Our benchmark (last row) is distinguished by explicitly incorporating reasoning traces, supporting a variety of question types, and covering a broader set of tasks and robotic platforms compared to prior work.}
    \resizebox{\textwidth}{!}{
    \begin{tabular}{lC{2cm}C{2cm}C{1cm}C{1.3cm}C{2.5cm}cc C{6cm} C{1.5cm}}
        \toprule
        \textbf{Benchmark} & \textbf{Modality} & \textbf{Question} & \textbf{Reasoning} & \textbf{\# of} & \textbf{Annotation} & \textbf{\# of}  & \textbf{\# of} & \textbf{Embodiments} & \textbf{\# of Robot} \\
        & & \textbf{Types} & \textbf{Trails} & \textbf{Questions} &  & \textbf{Tasks}  & \textbf{Videos} & & \textbf{Types}
        \\
        \midrule

        \rowcolor{LGray} Cosmos-R1 \cite{nvidia2025cosmosreason1physicalcommonsense} & Videos + Text & MCQ, TF & \textcolor{red}{\ding{55}} & 510 & Automatic + Human & 5 & 501 & Agibot-G1, Widow-X, UR5e,  Human Demonstrations & 2\\

        \rowcolor{LGray} Robo2VLM-1 \cite{chen2025robo2vlmvisualquestionanswering} & Images + Text & MCQ & \textcolor{red}{\ding{55}} & 6676 & Automatic & 5 & 463 &  Franka, UR5,fanuc-mate-200id, xArm, Kuka-LBR-IIWA & 1 \\

        \rowcolor{LGray} HRIBench \cite{shi2025hribenchbenchmarkingvisionlanguagemodels} & Image + Text &  Open & \textcolor{red}{\ding{55}} & 1000 & - & 5 & - & Quori, Kuri, Franka Emika & 3 \\

        \rowcolor{LGray} OpenEQA \cite{OpenEQA2023} &  Videos + Text & Open & \textcolor{red}{\ding{55}} & 1600  & Human Generated & 2 & 180 &  &  \\

        \rowcolor{LGray} PBench \cite{PBench_NVIDIA} & Image + Text & TF & \textcolor{red}{\ding{55}} & 5636 & Automatic + Human & 4 & 1044 & - & - \\

        \rowcolor{LGray} ECEBench \cite{dang2025ecbenchmultimodalfoundationmodels} & Video + Text & MCQ + Open & \textcolor{red}{\ding{55}} &  & Human Generated & 3 & 378 & - & - \\

        \rowcolor{violet!10}\textbf{Ours} & Videos/Image Frames + Text & TF + MCQ + Open & \textcolor{darkgreen}{\checkmark} & 1112 & Automatic + Human & 10 & 758 &  Agibot G1, Widow X, UR5e,  Human Demonstrations, Jackal, Hello  stretch, Franka & 3 \\
        \bottomrule
    \end{tabular}
    }
   
    
    \label{tab:methods_comparison}

%% file: sec/2_related_works.tex
\section{Related Work}
%
While Foundation Models have achieved remarkable fluency, establishing them with reasoning abilities, especially in spatial contexts, remains an actively researched challenge.

\noindent\textbf{Reasoning in Foundation Models.} Reasoning is a core component of intelligence, which is supported by the ability to formulate decisions, draw conclusions, and generalize conclusions in different domains. Recent Large Language Models (LLMs) and Large Multimodal Models (LMMs) have demonstrated remarkable improvements in factual recall and language generation\cite{openai2024gpt4technicalreport, chen2024internvl, bai2023qwenvlversatilevisionlanguagemodel}. Still, their performance in complex questions requiring multi-step reasoning and physical common sense falls short compared to human cognition. Recent breakthroughs have seen a rise in specialized reasoning models like GPT-o1 \cite{openai2024openaio1card} and Deepseek-R1 \cite{deepseekai2025deepseekr1incentivizingreasoningcapability}, which use deliberate thinking before coming to a conclusion. This marks a significant shift towards System 2 thinking \cite{kahnemanthinking}, where models take a deliberate, slow approach towards solving a problem.  Moreover, chain-of-thought prompting has been an effective methodology for LLMs to tackle complex questions by dividing the question into intermediate reasoning steps \cite{wei2023chainofthoughtpromptingelicitsreasoning}. 


\noindent\textbf{Physical Common Sense Reasoning.} 
Humans begin to learn about physical common sense within months after being born, and they start with weak, structured expectations about core concepts and refine them through passive observations and active interactions \cite{Baillargeon2004-os}. Through this, humans learn what is possible and impossible in a given scenario in the real world. With the widespread usage of LMMs, the need for physical common sense has become increasingly important. Benchmarks like PIQA test models on everyday physical common sense in text \cite{Bisk2020}. In the vision domain, benchmarks like PACS pioneered incorporating multisensory data for physical common-sense reasoning \cite{yu2022pacsdatasetphysicalaudiovisual}. More recently, Cosmos-R1 \cite{nvidia2025cosmosreason1physicalcommonsense} has emphasized grounding language models in physical world understanding. Cosmos-R1 introduces ontologies for space, time, and fundamental physics, and proposes a VQA benchmark to evaluate a model's grasp on the physical AI domain. 

\noindent\textbf{Reasoning Benchmarks.} Reasoning benchmarks are crucial for evaluating the logical skills of Large language models. CLEVR (Compositional Language
and Elementary Visual Reasoning) \cite{johnson2017clevr} is a synthetic benchmark explicitly designed as a diagnostic dataset that tests a range of visual reasoning abilities. These early benchmarks had a strong bias for superficial pattern matching, where a model could answer correctly without visual reasoning. To push beyond these limitations, benchmarks like Visual Common Sense Reasoning (VCR) \cite{zellers2019recognition} were introduced, where the model needs to answer correctly while providing a rational justification for the answer. More recently, multi-step reasoning benchmarks have emerged. Visual Reasoning Chain Benchmark (VRC-Bench) \cite{thawakar-etal-2025-llamav} assesses multi-step reasoning capabilities of models in 8 diverse categories, including math and medical imaging, and includes over 4000 human-verified reasoning steps. VRC-Bench also includes specific evaluation criteria to evaluate reasoning trails.

Physical Reasoning benchmarks have more recently been developed to test the models in physical common sense and embodied reasoning. Cosmos-R1 \cite{nvidia2025cosmosreason1physicalcommonsense} is a recent effort to measure physical common sense and embodied decision making by evaluating models using binary and multiple-choice questions. Even though Cosmos is a notable work in Physical reasoning benchmarks, it lacks the reasoning trails and any evaluation of reasoning trails. DriveLMM-o1 \cite{ishaq2025drivelmm} focuses on autonomous driving scenarios and provides over 22k visual QA examples. DriveLMM-o1 also proposes driving a specific evaluation metric to evaluate not only the answer but also the reasoning trails, to evaluate the quality of the reasoning in the same way as VRC-Bench.

%% file: sec/3_benchmark_curation.tex
\begin{figure}
    \centering
    \includegraphics[width=0.7\linewidth]{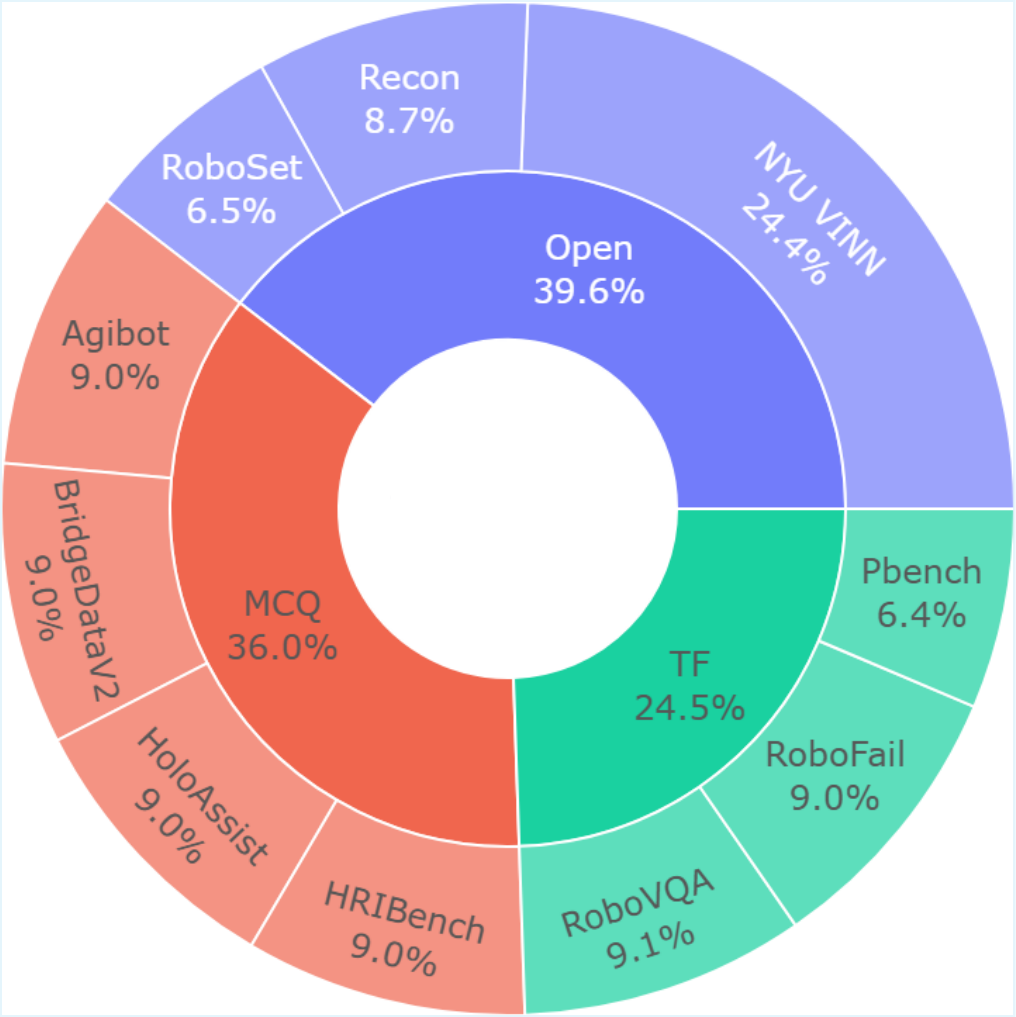}
    \caption{Dataset distribution and question type composition of our benchmark. Question types include open-ended questions (Open), multiple choice questions (MCQ), and True/False questions (TF). For clarity, we decompose the Cosmos-R1 benchmark into its constituent sub-datasets Agibot \cite{contributors2024agibotworldrepo}, BridgeDataV2 \cite{walke2024bridgedatav2datasetrobot}, HoloAssist \cite{HoloAssist2023}, RoboVQA \cite{robovqa2023arxiv}, and RoboFail \cite{liu2023reflect} to explicitly show the distribution of question types across these subsets.}
    \label{fig:datadistribution}
\end{figure}

\begin{figure*}[!htp]
    \centering
    \includegraphics[width=1\linewidth]{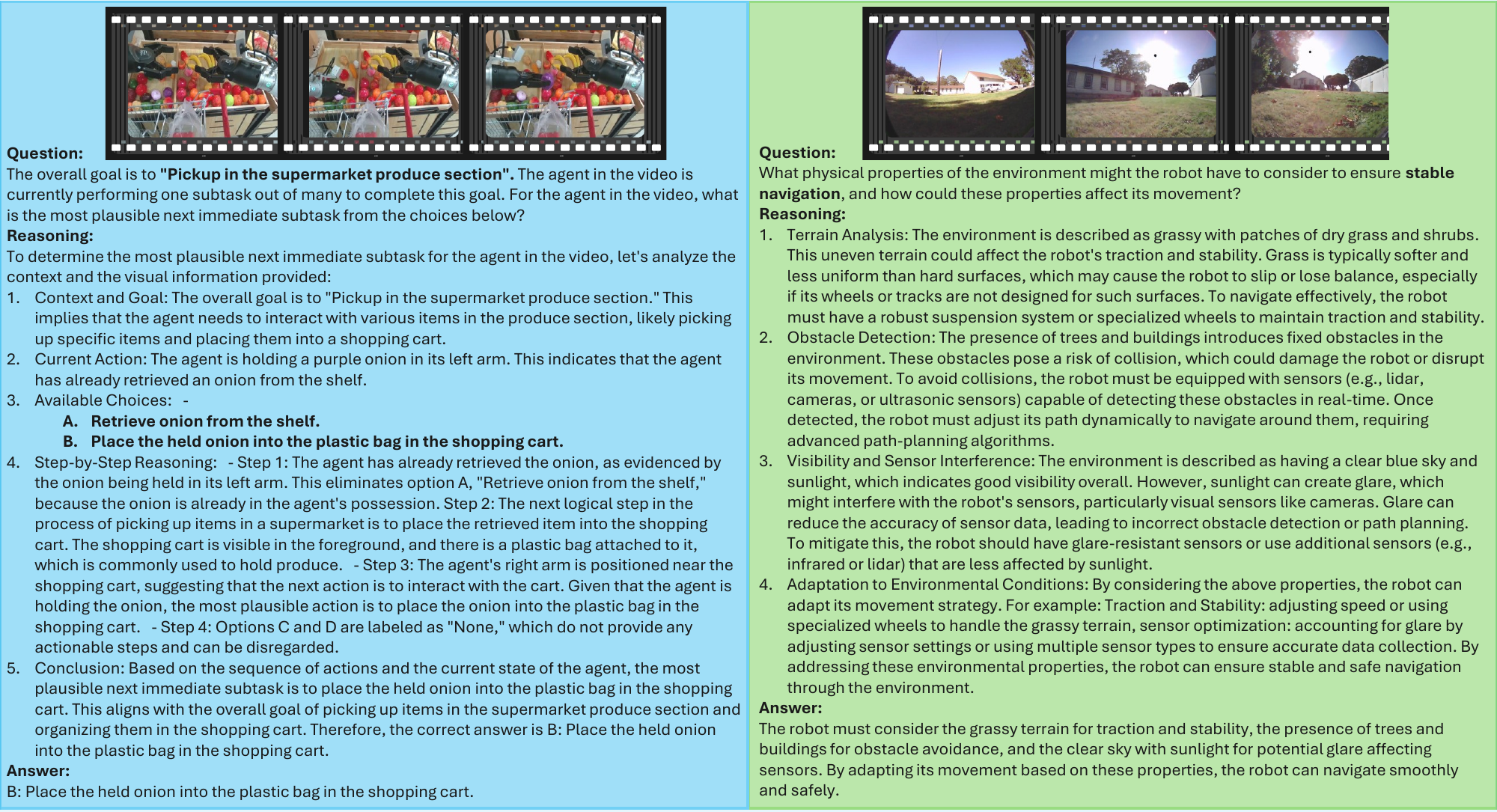}
    \caption{
    Two examples from our proposed benchmark. The left example involves a grasping task in a supermarket setting, where the agent must decide the most plausible next subtask based on the visual context and goal. The right example focuses on navigation, requiring the robot to reason about environmental properties like terrain and obstacles to ensure stable movement. Each example presents selected video frames, the question, the final answer, and detailed step-by-step reasoning that supports the decision.
    }
    \label{fig:qualitative}
\end{figure*}

\section{Benchmark Curation}

\begin{table}[t]
    \caption{Comprehensive evaluation criteria used to evaluate the quality of the reasoning done by models.}
    \centering
    \rowcolors{2}{white}{gray!6}
    \begin{tabularx}{\linewidth}{m{1.5cm}m{6cm}}
    \toprule
        Metric & Definition \\
        \midrule
        Faithfulness & Measures how well the reasoning steps in the LLM response align with the ground truth reasoning steps. \\
        \makecell[l]{Spatial\\Reasoning} & Measures the accuracy and quality of the reasoning in spatial tasks (e.g., object placement, navigation, coordinate systems). \\
        \makecell[l]{Physical\\Causality} & Evaluates the reasoning steps regarding the physical interactions or causes between objects, motions, forces, or processes. \\ 
        Safety & Assesses whether the reasoning process ensures safety in the robot's actions (e.g., collision avoidance, human interaction safety). \\
        Commonsense & Check for missing commonsense reasoning required to solve the problem in the robotics domain (e.g., understanding of basic physical principles, robot capabilities, environment).- Scoring Guidelines:   \\ 
        Hallucination & Detect irrelevant or invented reasoning steps not aligned with the source, particularly in robotics tasks. \\
        Redundancy & Identify redundant reasoning steps that do not add value to the robotics task. \\
        Semantic Coverage-Step & How well the reasoning covers the essential semantic elements of the task (e.g., environmental factors, object attributes, constraints). \\
        \makecell[l]{Reasoning\\Alignment} & Overall alignment between the hypothesis and the reference reasoning chain, taking robotics-specific constraints into account. \\
        Missing Step & Identify if any necessary reasoning steps are missing, particularly in robotics-specific processes. \\ 
        \bottomrule
    \end{tabularx}
    \label{tab:eval_metric}
\end{table}

We introduce the Foundation Model Embodied Reasoning benchmark (FoMER), designed to facilitate reasoning in physical AI scenarios. Our benchmark encompasses coverage for multiple robots across a broad range of robot modalities, enabling the assessment of LLM capabilities on diverse tasks such as next-action prediction, action affordance, physical common sense, temporal reasoning, tool use and manipulation, risk assessment, and robot navigation. 

\noindent\textbf{Benchmark Curation.} Given the broad scope of physical AI reasoning scenarios, we curate our benchmark from multiple datasets. Figure \ref{fig:datadistribution} shows the distribution of datasets that were used to curate the benchmark. Since we collect data from diverse sources covering different embodiments and tasks, we employ two different pipelines to create the necessary QA pairs and reasoning trails. 

\noindent\textbf{QA and reasoning trail generation pipeline. }  
Among the datasets we selected, the NYU VINN \cite{pari2021surprisingeffectivenessrepresentationlearning}, Recon \cite{shah2021rapid}, and Roboset \cite{kumar2023robohiveunifiedframeworkrobot} datasets did not include question-answer pairs. Therefore, we generated QA pairs along with corresponding reasoning trails for these datasets.
To ensure we have a wide variety of question types, we choose to curate open-ended questions from these datasets. To guarantee that QA pairs address challenging reasoning skills (e.g. physical common sense, spatial and temporal reasoning, tool use and manipulation, and risk assessment), we first prompted Qwen2.5-VL-32B-Instruct \cite{Qwen2.5-VL} to generate an exhaustive inventory of all visible objects in the scenario and to describe dynamic elements and interactions occurring in the scene \cite{nvidia2025cosmosreason1physicalcommonsense}.  Following this, we prompted Qwen once more to generate QA pairs with chain-of-thought explanations based on the previously extracted objects and actions, and the visual scene. Using this method, we were able to generate challenging QA pairs and reasoning trails of the answers. 
For the datasets already containing QA pairs, we instead prompted Qwen to generate reasoning trails that logically lead to the final answer by providing the visual information and the QA pairs. 

\noindent\textbf{Manual Verification.} Considering LLMs can sometimes generate erroneous responses \cite{10.1145/3703155}, a manual verification of the Q\&As and reasoning trails was needed to ensure the quality of the benchmark. 
The verification process involved a thorough review of the generated QA pairs and reasoning trails.
We asked the verifiers to check whether the questions are relevant, physically plausible, and aligned with our intended task categories. For the reasoning trails, we instructed the verifiers to add or remove steps that were not necessary. Finally, approximately 12\% of questions were removed because they were not aligned with our tasks or deemed too trivial.  
Figure \ref{fig:qualitative} provides qualitative examples of the questions, answers, and reasoning trails included in our benchmark.

\noindent\textbf{Task Ontology. } To enable fine-grained evaluation of physical reasoning abilities, we categorize the questions into ten diverse task types: Task completion verification, next-action prediction, action affordance, physical common sense reasoning, robot-centric, temporal reasoning, tool use and manipulation, social navigation, human-robot object interaction, and risk assessment. 
Each task type is designed to test specific aspects of physical reasoning, ensuring a comprehensive evaluation.


%% file: sec/4_evaluation_framework.tex

\section{Evaluation Framework}
To effectively evaluate the embodied reasoning capabilities of LMMs, we propose a structured evaluation framework that goes beyond simple accuracy metrics.
We leverage an LLM-as-judge approach for automated evaluation, with both closed-source (GPT-4o \cite{openai2024gpt4technicalreport}) and open-source models (Qwen3-32B \cite{qwen3technicalreport}).
While it is important to measure whether a model reaches or selects the correct action in a given scenario, it is equally critical to assess why the model arrived at that decision, particularly in embodied contexts where safety, physical feasibility, and spatial awareness are essential.
Our benchmark is built around scenarios that demand more than recognition or memorization; they require coherent, context-aware reasoning grounded in visual input, task constraints, and the physical properties of the environment. 

For each example in our benchmark, we make use of our evaluators to score both the final response and the reasoning steps. 
These evaluators are guided by a comprehensive rubric with ten clearly defined criteria, each scored on a 1–10 scale. The criteria include Faithfulness, Spatial Accuracy, Physical Plausibility, Safety, Commonsense, Hallucination, Redundancy, Semantic Coverage, Completeness, and Answer Alignment, as described in the Table \ref{tab:eval_metric}. Together, these dimensions capture both the correctness and the quality of a model's reasoning in embodied settings.
To ensure consistency and minimize evaluation bias, we use a prompt template that explicitly instructs the evaluator to compare the model’s prediction with both the original question and a human-annotated ground truth answer. Each metric is scored according to a well-defined rubric with granular thresholds, and the results are returned in a standardized JSON format suitable for integration with our benchmarking pipeline.
This setup allows us to analyze not just whether a model reaches the right answer, but whether it uses the right reasons. Otherwise, hallucinated assumptions or unsafe recommendations might go undetected.

By averaging across all metrics, we produce an overall reasoning accuracy that reflects the model's capacity to make grounded, context-sensitive decisions. As we demonstrate in our experiments, this multi-dimensional evaluation reveals significant variation in performance across current LMMs, highlighting specific areas, such as human-robot-object interaction or social navigation, where progress is still needed.


Our benchmark includes three question types: multiple-choice (MCQ), true/false (TF), and open-ended questions.
We make use of two distinct prompts: one for evaluating open-ended questions and another for MCQs and TF questions.
For MCQ and TF questions, evaluators were provided with the question, answer, and options, and were asked to assess the final accuracy, assigning a score of either 0 or 10 \cite{liu2023visual}. Evaluating open-ended questions was more challenging, as there can be multiple partially correct answers in the embodied context. To ensure fair evaluation, we applied a subset of the criteria used for evaluating reasoning trails. Evaluators were guided by factors including factual accuracy, physical plausibility, temporal consistency, redundancy, commonsense, and safety awareness, and asked to assign an overall score between 0 and 10. By employing these two complementary evaluation methods, we were able to assess the final answer accuracy more fairly and consistently across all question types.

%% file: sec/5_benchmarking.tex
\section{Benchmarking}

\begin{table*}[!htp] 
\input{Tables/DatasetwiseResults}

\end{table*}

In our benchmark, we employ two prompting strategies depending on the question type.
For MCQ and TF questions, the models were provided with the visual context, question, and options, and were instructed to provide an answer selected from the given options alongside a detailed step-by-step explanation that logically leads to the final answer. As for open-ended question, we prompted the models with the visual context and the question and asked them to provide a detailed step-by-step explanation with the final answer. Given that different models have different capabilities when it comes to prompting with visual frames, and considering most of our videos are short (1-10 seconds), we chose to prompt all the models with 8 uniformly sampled frames from the videos. The only exception is the Robofail subset from Cosmos-R1, which contains long videos from 3 to 5 minutes; for these, we sampled 32 uniformly spaced frames. NYU VINN \cite{pari2021surprisingeffectivenessrepresentationlearning}, Recon \cite{shah2021rapid}, and Roboset \cite{kumar2023robohiveunifiedframeworkrobot} consisted of static image frames rather than videos, and we prompted these subsets with 8 sampled images. As for Pbench and HRIBench, which have only one image, we prompted the models with the single image \cite{zhang2024video}. To ensure fairness, we standardize prompts across all models according to question type and restricted the output length to 4096 tokens (or the closest applicable limit per model).

\begin{figure}
    \centering
    \includegraphics[width=1\linewidth]{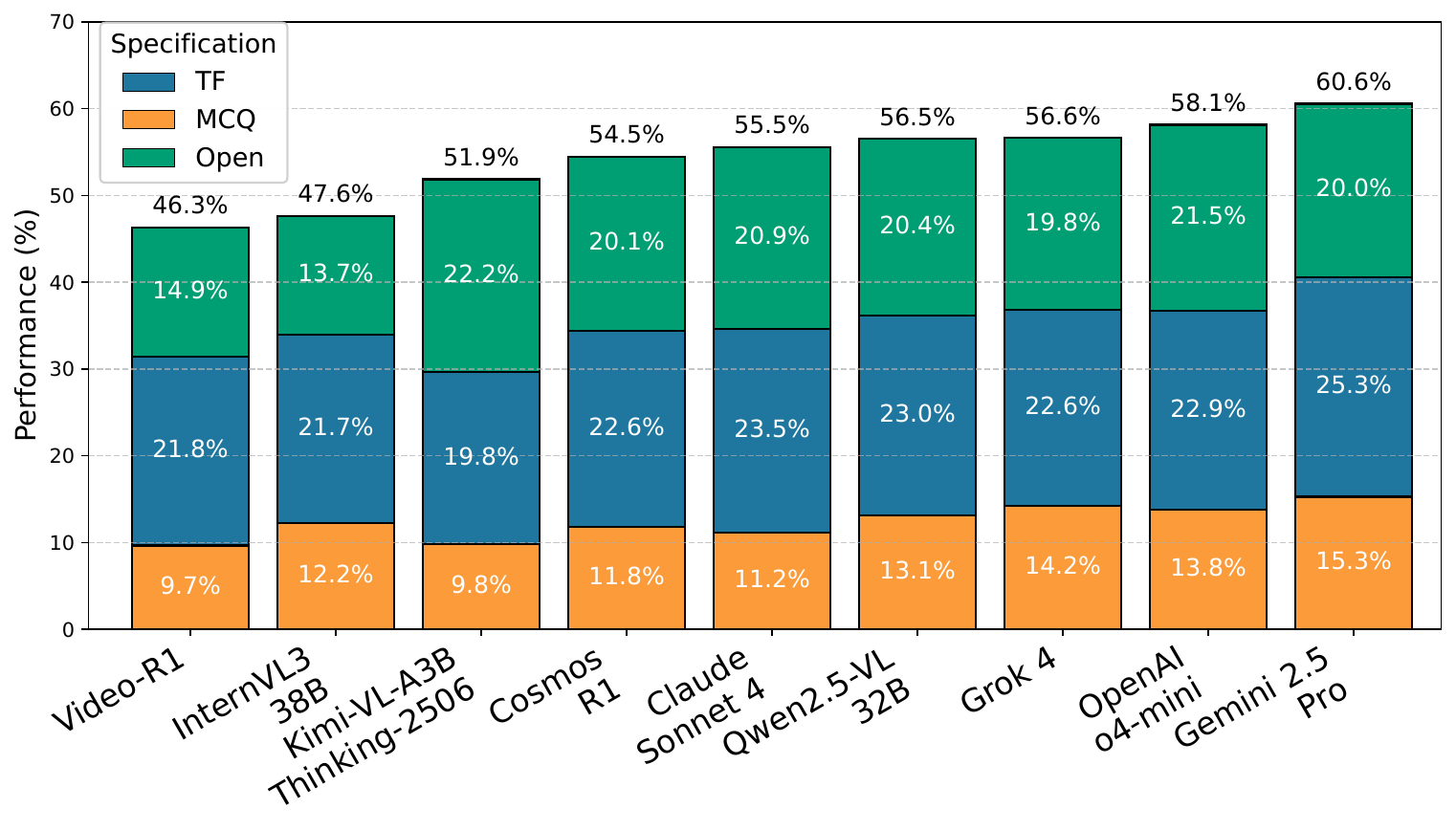}
    \caption{Comparison of the final accuracies achieved by different models in each question type of the benchmark. Considering the results, models perform better on TF questions and open-ended questions compared to MCQ questions. }
    \label{fig:qtypewise}
\end{figure}

In the evaluation framework, the GPT-4o model was used as a judge to evaluate both the final answers and the reasoning trails. To ensure reproducibility, we fixed the evaluation model to the "gpt-4o-2024-05-13" variant, ensuring that results remain stable across different runs and future model updates. Additionally, we experimented with using the open-source Qwen3-32B \cite{qwen3technicalreport} model as a judge, which yielded consistent results, as shown in Table \ref{tab:acc_model}. This demonstrates that our evaluation framework can be generalized across different judging LLMs.

\begin{figure}
    \centering
    \includegraphics[width=1\linewidth]{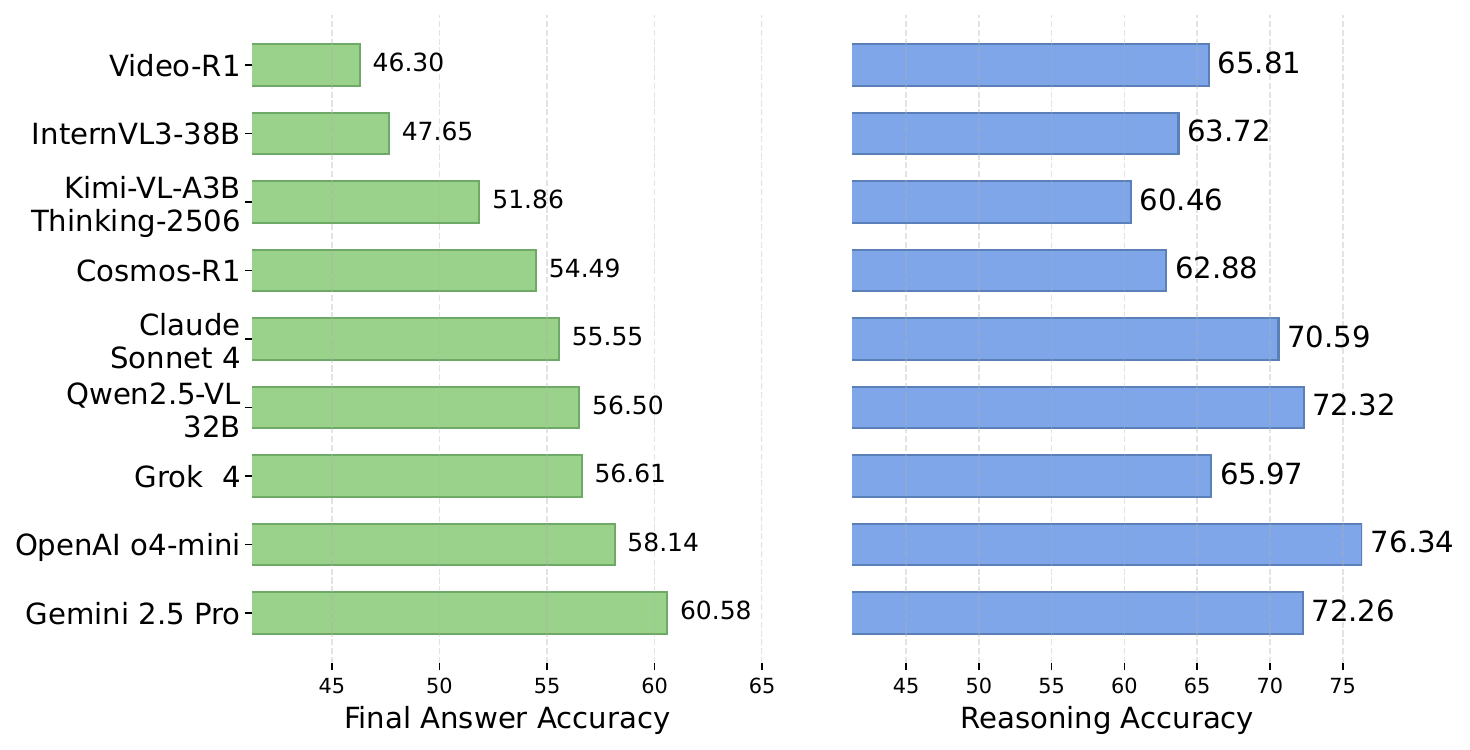}
    \caption{Performance of different open source as well as closed source SoTA models, highlighting the reasoning accuracy as well as the final accuracy. Here, we evaluate the reasoning steps thoroughly using our proposed evaluation criteria.}
    \label{fig:overall_results}
\end{figure}

Figure \ref{fig:overall_results} presents the overall average reasoning accuracy and the overall final accuracy score of each model. The results offer interesting insights: Closed-source models like Gemini 2.5 Pro \cite{comanici2025gemini25}, OpenAI o4-mini \cite{openai2025o4mini_systemcard}, Grok 4 \cite{xai2025grok4docs}, and Claude Sonnet 4 \cite{anthropic2025claude4systemcard} models generally perform better compared to the open-source models, and the only exception is Qwen2.5-VL-32B model, which performs comparably with Grok 4 and Claude Sonnet 4 models. OpenAI o4-mini shows the best generalized reasoning capabilities among the models. Video-R1 \cite{feng2025video} is the lowest performing model, followed by InternVL3-38B \cite{chen2024internvl}, and Kimi-VL-A3B-Thinking-2506 \cite{kimiteam2025kimivltechnicalreport}.  As for the reasoning accuracy evaluations, we measure the pure reasoning and generalization capabilities of the models; hence, we did not consider the final answer correctness when evaluating the reasoning accuracy. As a result, some models with lower performance, such as Video-R1\cite{feng2025video}, were able to achieve good reasoning accuracies. This was especially necessary since the main goal of evaluating reasoning trails was to evaluate the model's capability of producing well-grounded and context-sensitive decisions. 
This helps identify models that understand the problem but fail in execution, versus those that succeed through guessing or pattern exploitation.

    

\begin{table}[]
    \centering
    \caption{Performance of Gemini 2.5 Pro when prompted with 8 frames, prompted with the whole video as the visual context, and prompted with only the middle image. Results are reported on the subset of our benchmark where video-based visual context is available.}
    \begin{tabular}{l|c|c|c}
    \toprule
        Model \textbackslash Setting & 8 frames & With Video & Middle Frame \\
                \midrule
        Gemini 2.5 Pro & 62.97\% & 69.62\% & 51.89\%\\
        \bottomrule
    \end{tabular}
    
    \label{tab:visual_type}
\end{table}

\begin{figure}
    \centering
    \includegraphics[width=1\linewidth]{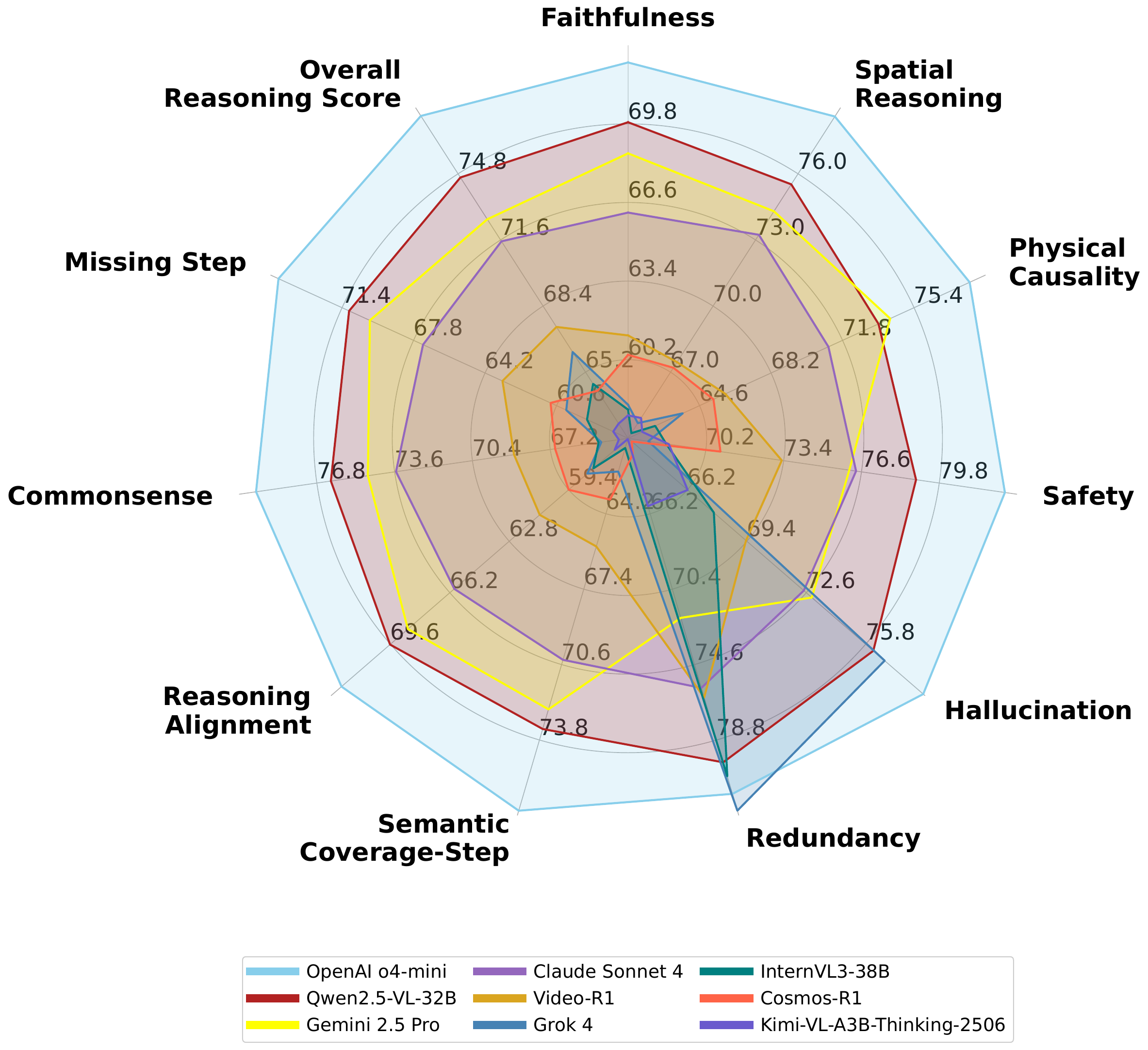}
    \caption{A breakdown of the model performance on the proposed evaluation criteria, where we evaluate the reasoning traces of the models in physical AI-specific criteria like spatial reasoning, physical causality, commonsense, and safety.}
    \label{fig:radar}
\end{figure}

Figure \ref{fig:radar} shows the reasoning accuracies of the models in our different evaluation criteria. OpenAI o4-mini gives the highest average accuracy while performing consistently in all of our evaluation criteria. Qwen, Gemini, and Claude also perform consistently in all criteria. Surprisingly, Cosmos-R1 does not perform well in any of our reasoning criteria, even though it achieves a good 56.83\% final answer accuracy.  

We also discuss question type-wise performance on our dataset in Figure \ref{fig:qtypewise}. Overall, models perform best on TF questions compared to MCQ and open-ended questions. 
This is primarily because many of the TF questions correspond to task completion verification. 
The models performed worse on MCQ questions and moderately on open-ended questions in our benchmark. The same trend holds consistently across all the models. We extensively analyze the performance of each model on our benchmark in Table \ref{tab:dataset_wise}. Among them, RoboVQA, Pbench, and RoboFail, which contain TF questions, show higher model performance. Recon, NYU VINN, and RoboSet consist of open-ended questions, where models also perform relatively well. In contrast, sub-datasets comprising MCQs such as Agibot, HoloAssist, BridgeDataV2, and HRIBench consistently yield lower performance across models.

\begin{figure}
    \centering
    \includegraphics[width=1\linewidth]{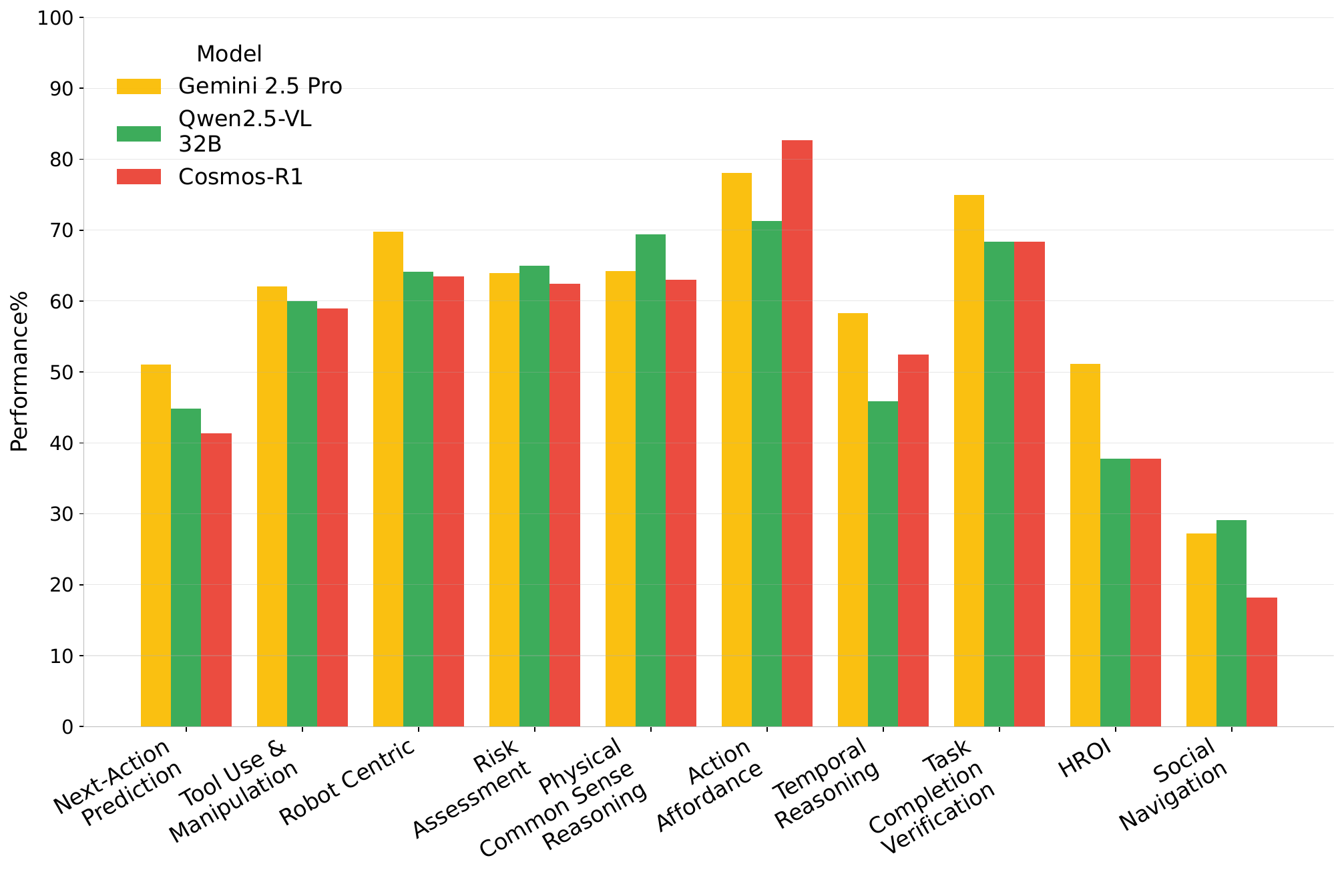}
    \caption{Performance comparison of Gemini 2.5 Pro, Qwen2.5-VL 32B, and Cosmos-R1 on different categories of our benchmark. The results highlight that action affordance is comparatively easier for models, whereas social navigation is the hardest.}
    \label{fig:categorywise}
\end{figure}

We further analyze performance across task categories in Figure~\ref{fig:categorywise}. Gemini 2.5 Pro outperforms Qwen2.5-VL-32B and Cosmos-R1 in all categories except for action affordance, risk assessment, physical common sense reasoning, and social navigation. Qwen2.5-VL-32B achieves the best results on physical common sense reasoning, risk assessment, and social navigation, while Cosmos-R1 attains the highest score on action affordance, exceeding 80\%. Interestingly, all models perform poorly on human–robot object interaction (HROI) and social navigation, both of which demand deeper reasoning about social norms and conventions in order to answer precisely.

Considering a subset of our benchmark contains videos instead of static image frames, we use Gemini 2.5 Pro to test the performance differences of the models when prompted with the whole video and when prompted with some selected frames, and the results are presented in Table \ref{tab:visual_type}. As expected, Gemini seems to perform better when prompted with the video file, achieving 69.62\% when passing the video file due to the model's ability to analyze the scene as a cohesive spatio-temporal volume and analyze the motion flow between frames, unlike when prompting with sampled frames. To show the need for the temporal context to answer the questions, we prompted the same model with only the middle image, and the performance dropped by a significant amount from our 8-frame baseline from 62.97\% to 51.89\%, emphasizing the importance of the temporal context in physical AI reasoning scenarios.

\begin{table}[t]
    \centering
    \caption{To ensure the fairness of our automatic evaluation system, which uses GPT-4o as a judge, we conducted a human evaluation study. Three volunteers were asked to answer a subset of benchmark questions covering all three question categories (MCQ, TF, and open-ended). The results of this experiment are presented here.}
    \begin{tabular}{l|c|c}
    \toprule
         & GPT-4o  & Human\\
         & as judge (\%) & as judge (\%)\\
        \midrule
        Human Baseline & 80.93 & 84.47 $\pm$ 0.63 \\
        Gemini 2.5 Pro & 60.02 & 65.00 $\pm$ 0.31\\
        \bottomrule
    \end{tabular}
    \label{tab:human_analysys}
\end{table}

To ensure the fairness of our automatic evaluation system \cite{chen2024humans}, which uses an LLM as a judge, we conducted a human evaluation study. Three volunteers were asked to answer a subset of 50 benchmark questions covering all three question categories (MCQ, TF, and open-ended). Their responses were then evaluated using our automatic system (GPT-4o as judge), and the volunteers also scored each other’s answers for comparison. In addition, we asked the volunteers to evaluate the responses of Gemini 2.5 Pro on the same set of questions. Table \ref{tab:human_analysys} presents the findings, reporting average scores across the three human evaluators. We observe that both humans and GPT-4o assigned similar scores to the human responses and to Gemini 2.5 Pro’s answers, supporting the reliability of our automatic evaluation system for benchmarking and model comparison. Moreover, this result indicates that humans were able to score significantly higher (around 30\%) compared to the highest performing model (Gemini 2.5 Pro).

\begin{table}
\centering
\caption{Comparison of reasoning and final accuracy scores evaluated by GPT-4o and Qwen3-32B across selected models.}
\setlength{\tabcolsep}{4pt}            
\renewcommand{\arraystretch}{1.15}
\resizebox{\columnwidth}{!}{           
\begin{tabular}{lcccc}
\toprule
\multicolumn{1}{c}{\textbf{Model}} &
\multicolumn{2}{c}{\textbf{GPT-4o}} &
\multicolumn{2}{c}{\textbf{Qwen3-32B}}\\
\cmidrule(lr){2-3}\cmidrule(lr){4-5}
& \makecell{Reasoning\\Accuracy} & \makecell{Final\\Accuracy}
& \makecell{Reasoning\\Accuracy} & \makecell{Final\\Accuracy} \\
\midrule
OpenAI o4-mini       & 76.34 & 58.14 & 76.11    & 61.62 \\
Cosmos-Reason1-7B & 62.88 & 54.49 & 64.03 & 54.66 \\
Gemini 2.5 Pro    & 72.26 & 60.58 &   71.88   &  62.39    \\
\bottomrule
\end{tabular}}
\label{tab:acc_model}
\end{table}

%% file: Tables/DatasetwiseResults.tex
\centering
\caption{
Comparison of reasoning accuracy (RA) and final accuracy (FA) across datasets of our benchmark, including Cosmos-R1 (decomposed into Agibot, BridgeDataV2, HoloAssist, RoboVQA, and RoboFail), RoboSet, NYU VINN, Recon, HRIBench, and Pbench.}
\setlength{\tabcolsep}{4pt}            
\renewcommand{\arraystretch}{1.15}
\resizebox{\textwidth}{!}{           
\begin{tabular}{lcc|cc|cc|cc|cc|cc|cc|cc|cc|cc}
\toprule
&
\multicolumn{2}{c}{\textbf{RoboVQA}} &
\multicolumn{2}{c}{\textbf{Agibot}} & 
\multicolumn{2}{c}{\textbf{Robofail}} &
\multicolumn{2}{c}{\textbf{HoloAssist}} &
\multicolumn{2}{c}{\textbf{BridgeDataV2}} &
\multicolumn{2}{c}{\textbf{Pbench}} &
\multicolumn{2}{c}{\textbf{RoboSet}} &
\multicolumn{2}{c}{\textbf{Recon}} &
\multicolumn{2}{c}{\textbf{NYU VINN}} &
\multicolumn{2}{c}{\textbf{HRIBench}} \\
\cmidrule(lr){2-3}\cmidrule(lr){4-5}\cmidrule(lr){6-7}\cmidrule(lr){8-9}\cmidrule(lr){10-11}\cmidrule(lr){12-13}\cmidrule(lr){14-15}\cmidrule(lr){16-17}\cmidrule(lr){18-19}\cmidrule(lr){20-21}
\multicolumn{1}{c}{\textbf{Model}} & \makecell{RA} & \makecell{FA}
& \makecell{RA} & \makecell{FA}
& \makecell{RA} & \makecell{FA}
& \makecell{RA} & \makecell{FA}
& \makecell{RA} & \makecell{FA}
& \makecell{RA} & \makecell{FA}
& \makecell{RA} & \makecell{FA}
& \makecell{RA} & \makecell{FA}
& \makecell{RA} & \makecell{FA}
& \makecell{RA} & \makecell{FA} \\
\midrule
\rowcolor{LGray1}

\rowcolor{LGray1}
Video-R1 \cite{feng2025video}	&73.6	&80.2	&61.2	&36.0	&67.1	&58.0	&68.8	&32.0	&60.3	&28.0 &67.1	&67.6	&63.2	&32.7	&73.0	&53.3	&72.2	&48.3	&51.6	&26.0 \\
InternVL3-38B \cite{chen2024internvl}	&74.6	&72.3	&67.2	&42.0	&69.8	&\underline{62.0}	&72.8	&\underline{56.0}	&65.8	&33.0	&26.2	&71.8	&\underline{69.3}	&30.4	&70.6	&56.2	&69.3	&41.9	&55.5	&23.0 \\
\rowcolor{LGray1}
Kimi-VL-A3B \cite{kimiteam2025kimivltechnicalreport}	&65.9	&66.3	&59.3	&30.0	&61.9	&52.0	&62.6	&39.0	&56.4	&29.0	&56.9	&63.4	&59.9	&58.4	&70.5	&71.0	&\underline{71.0}	&\underline{68.6}	&40.2	&21.0 \\
Cosmos-R1 \cite{nvidia2025cosmosreason1physicalcommonsense}	&79.1	&\textbf{85.2}	&61.1	&45.0	&67.2	&57.0	&63.6	&46	&52.5	&27.0	&61.8	&60.0	&57.0	&53.5	&71.8	&66.8	&64.7	&62.4	&50.2	&27.0 \\
\rowcolor{LGray1}
Claude Sonnet 4 \cite{anthropic2025claude4systemcard}	&78.3	&81.2	&68.2	&34.0	&70.8	&58.0	&73.6	&43.0	&\underline{70.4}	&32.0	&74.8	&80.3	&62.4	&41.3	&72.0	&68.8	&73.3	&\textbf{69.2}	&62.1	&29.0 \\
Qwen2.5-VL-32B \cite{Qwen2.5-VL}	&\textbf{83.8}	&72.3	&70.4	&40.0	&73.5	&60.0	&71.8	&52.0	&69.7	&\underline{36.0}	&\underline{76.9}	&\underline{83.1}	&69.2	&47.6	&\underline{76.7}	&70.8	&\textbf{79.6}	&63.6	&51.6	&33.0 \\
\rowcolor{LGray1}
Grok 4 \cite{xai2025grok4docs}	&72.3	&63.4	&71.3	&34.0	&70.6	&59.0	&75.8	&36.0	&66.5	&\textbf{36.5}	&76.8	&71.8	&68.7	&53.8	&73.7	&69.0	&69.1	&55.7	&64.1	&35.0 \\
OpenAI o4-mini \cite{openai2025o4mini_systemcard}	&79.8	&70.3	&\underline{74.2}	&\textbf{53.0}	&\textbf{77.3}	&61.0	&\textbf{80.0}	&44.0	&\textbf{71.5}	&33.0	&\textbf{79.3}	&\textbf{84.5}	&\textbf{75.3}	&\textbf{63.2}	&\textbf{82.1}	&\textbf{72.0}	&\underline{78.1}	&64.8	&\underline{65.8}	&\textbf{40.0} \\
\rowcolor{LGray1}
Gemini 2.5 Pro \cite{comanici2025gemini25}	&\underline{82.7}	&\underline{83.2}	&\textbf{75.0}	&\underline{52.0}	&\underline{73.9}	&\textbf{73.0}	&\underline{79.9}	&\textbf{70.0}	&70.3	&29.0	&68.7	&78.9	&66.6	&\underline{61.7}	&74.1	&70.3	&62.2	&58.7	&\textbf{69.2}	&\underline{38.0} \\

\bottomrule
\end{tabular}}
\label{tab:dataset_wise}

%% file: sec/6_conclution.tex
\section{CONCLUSION}

In this paper, we introduced the Foundation Model Embodied Reasoning(FoMER) benchmark designed to evaluate the embodied reasoning capabilities of LLMs. The benchmark consists of over 1,000 samples with detailed reasoning steps spanning 10 diverse task categories. In addition, we proposed a new evaluation framework that jointly assesses both action validity and reasoning correctness. We analyzed the performance of nine state-of-the-art models, including both open-source and proprietary systems. Our results reveal significant limitations of current models on embodied reasoning tasks and underscore the importance of analyzing and evaluating reasoning trails to better understand model capabilities. As a result, FoMER could serve as a testbed to identify potentially unsafe or unreliable reasoning in LLMs and agentic models before real-world deployment.